\documentclass[lettersize,journal]{IEEEtran}
\usepackage{amsmath,amsfonts}
\usepackage{algorithmic}
\usepackage{algorithm}
\usepackage{array}
\usepackage[caption=false,font=normalsize,labelfont=sf,textfont=sf]{subfig}
\usepackage{textcomp}
\usepackage{stfloats}
\usepackage{url}
\usepackage{verbatim}
\usepackage{graphicx}
\usepackage{cite}

\usepackage{amsmath,amssymb,amsfonts}
\usepackage{stfloats}
\usepackage{algorithmic}
\usepackage{graphicx}
\usepackage{textcomp}
\usepackage{xcolor}
\usepackage{multirow}
\usepackage{makecell}
\usepackage{booktabs}
\usepackage{subcaption}
\usepackage{bm}
\usepackage{bbding}
\usepackage[colorlinks,
            linkcolor=red,  
            anchorcolor=blue,
            citecolor=black,  
            ]{hyperref}

\let\oldtwocolumn\twocolumn
\renewcommand\twocolumn[1][]{%
    \oldtwocolumn[{#1}{
    \begin{center}
           \includegraphics[width=0.98\textwidth]{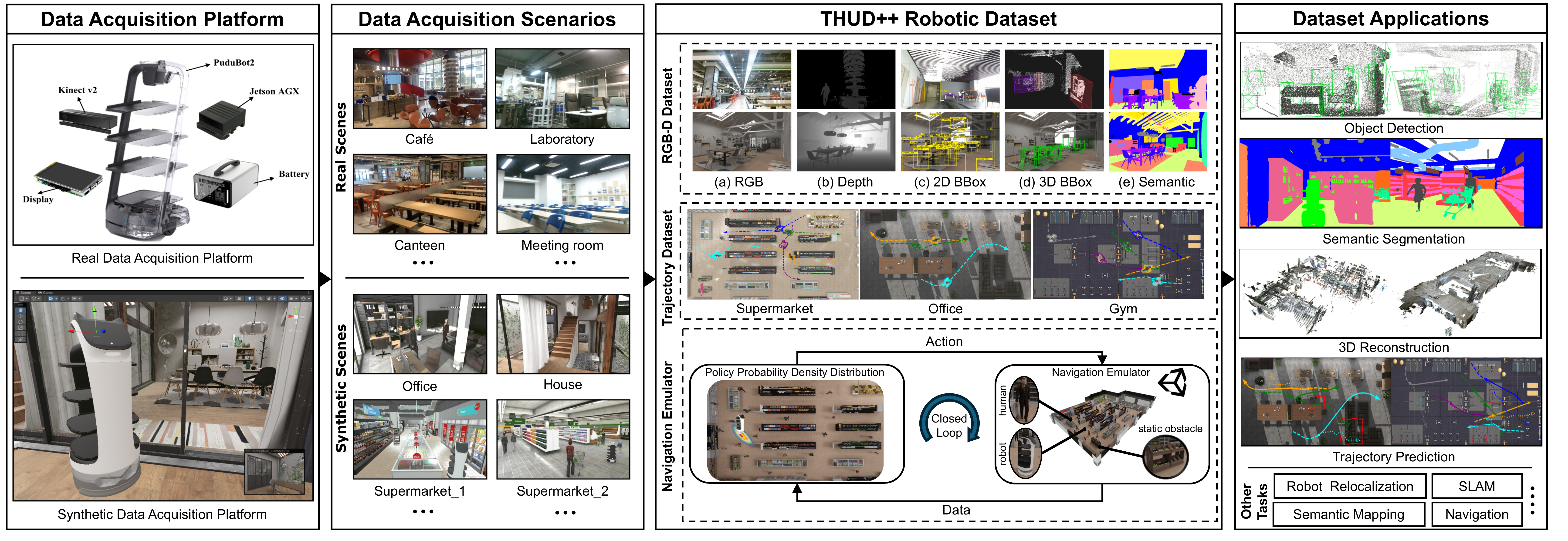}
           \captionof{figure}{THUD++ robotic dataset, first column: real and synthetic data acquisition platforms; second column: real and synthetic scenarios; third column: dataset components and annotations; fourth column: supported applications.}
           \label{fig0}
        \end{center}
    }]
}

\begin{document}

\title{THUD++: Large-Scale Dynamic Indoor Scene Dataset and Benchmark for Mobile Robots}

\author{Zeshun Li, Fuhao Li, Wanting Zhang, Zijie Zheng, Xueping Liu,

Yongjin Liu,~\IEEEmembership{Senior Member,~IEEE}, and Long Zeng,~\IEEEmembership{Member,~IEEE}
\thanks{Manuscript created December, 2024. (\textit{Zeshun Li, Fuhao Li and Wanting Zhang contributed equally to this work.}) (\textit{Corresponding author: Long Zeng.})}
\thanks{Zeshun Li, Fuhao Li, Wanting Zhang, Zijie Zheng, Xueping Liu and Long Zeng are with the Department of Advanced Manufacturing, Shenzhen International Graduate School, Tsinghua University, Shenzhen 518000, China. (e-mail: \{lzs23, lfh23, zhangwt23, zhengzj22\}@mails.tsinghua.edu.cn, \{liuxp, zenglong\}@sz.tsinghua.edu.cn)}

\thanks{Yongjin Liu is with BNRist, MOE Key Laboratory of Pervasive Computing, Department of Computer Science and Technology, Tsinghua University, Beijing 100084, China. (e-mail: liuyongjin@tsinghua.edu.cn).}

}

\markboth{}%
{Li \MakeLowercase{\textit{et al.}}: THUD++}


\maketitle

\begin{abstract}
Most existing mobile robotic datasets primarily capture static scenes, limiting their utility for evaluating robotic performance in dynamic environments. To address this, we present a mobile robot oriented large-scale indoor dataset, denoted as THUD++ (TsingHua University Dynamic) robotic dataset, for dynamic scene understanding. 
Our current dataset includes 13 large-scale dynamic scenarios, combining both real-world and synthetic data collected with a real robot platform and a physical simulation platform, respectively. The RGB-D dataset comprises over 90K image frames, 20M 2D/3D bounding boxes of static and dynamic objects, camera poses, and IMU. The trajectory dataset covers over 6,000 pedestrian trajectories in indoor scenes. Additionally, the dataset is augmented with a Unity3D-based simulation platform, allowing researchers to create custom scenes and test algorithms in a controlled environment.
We evaluate state-of-the-art methods on THUD++ across mainstream indoor scene understanding tasks, e.g., 3D object detection, semantic segmentation, relocalization, pedestrian trajectory prediction, and navigation. Our experiments highlight the challenges mobile robots encounter in indoor environments, especially when navigating in complex, crowded, and dynamic scenes. By sharing this dataset, we aim to accelerate the development and testing of mobile robot algorithms, contributing to real-world robotic applications. The dataset is shared at: \href{https://jackyzengl.github.io/THUD-plus-plus.github.io/}{https://jackyzengl.github.io/THUD-plus-plus.github.io/.}

\end{abstract}


\section{Introduction}
\IEEEPARstart{M}{obile} robots are increasingly deployed in diverse indoor settings, such as restaurants and supermarkets, where they operate in highly dynamic environments characterized by numerous moving objects and people. Navigating in cluttered and unpredictable environments demands robust perception, prediction and planning systems capable of handling frequent occlusions, unexpected obstacles and complex human-robot interactions \cite{hu2023planning}. Additionally, ensuring safe and reliable operation in shared human-robot spaces requires social awareness, and adherence to human behavioral norms \cite{kruse2013human, du2020online}. In order to enable mobile robots to better handle complex dynamic scenes, high-quality datasets are indispensable.

However, most existing mobile robot datasets primarily focus on static scenes \cite{B3DO, NYU, SUN3D, Stanford, SceneNet, SUNRGBD, ScanNet, SUNCG, Matterport3D, InteriorNet, ARKitScenes, ScanNet++}, limiting their effectiveness in training and evaluating robots under practical working conditions, especially in large-scale dynamic indoor environments. This limitation is critical, as static environments fail to capture the complexity and unpredictability of real-world indoor settings. To bridge this gap, we introduce our mobile robot oriented large-scale indoor dataset named THUD++ (TsingHua University Dynamic) dataset, specifically designed to support mobile robot training and evaluation across diverse tasks in dynamic indoor environments. An overview of THUD++ is shown in  Fig.~\ref{fig0}. 

THUD++ comprises three primary components: an RGB-D dataset, a pedestrian trajectory dataset, and a robot navigation emulator. It integrates both real-world and synthetic scene data. Real-world data was captured using a mobile robot, while synthetic scene data was generated via our Unity3D-based simulation platform. \textit{The RGB-D dataset} is collected from 13 dynamic scenarios (8 real-world and 5 synthetic scenes), each exhibiting different levels of dynamic complexity. This dataset is densely annotated with ground-truth labels, including over 90,000 frames (each containing an RGB image and a depth map), more than 20 million 2D/3D object detection bounding boxes, semantic segmentation annotations, and camera poses. \textit{The trajectory prediction dataset} is collected from 3 synthetic indoor scenes. These scenes are designed to capture realistic pedestrian movement patterns in confined spaces, with dynamic elements such as moving pedestrians incorporated to simulate typical indoor traffic. This dataset contains over 6,000 frames spanning approximately 60 minutes of recorded data. \textit{The robot navigation emulator} contains over 10 scenes, each constructed to simulate real-world environments. The scenes were meticulously designed to encompass diverse layouts. Dynamic scenarios were generated by populating these scenes with pedestrians, including running children, people pushing shopping carts, and other moving robots.

We evaluated and tested THUD++ using representative algorithms for both static and dynamic mobile robot indoor tasks, including 3D object detection \cite{10373157, wang2023multi, mao20233d}, semantic segmentation \cite{muhammad2022vision, yuan2024survey, li2024transformer}, robot relocalization \cite{miao2024survey}, trajectory prediction \cite{fang2024behavioral}, and navigation \cite{singamaneni2024survey, mirsky2024conflict}. The results show that algorithms designed for different tasks experience varying levels of performance degradation in scenes with dynamic objects.

\textbf{Contributions and advantages of THUD++.} By utilizing the THUD++ dataset, we aim to provide a valuable resource for researchers in mobile robotics, advancing and optimizing mobile robot systems. By making this dataset publicly available, we aim to engage a broader research community, fostering continuous innovation and progress in mobile robotics research. In this work, we make four main contributions:
\begin{itemize}
\item We provide an RGB-D dataset annotated with dynamic instances for large-scale indoor scenes, containing both real and synthetic data that closely resemble the operational environments of mobile robots. This dataset poses significant challenges for dynamic robot tasks.
\item We provide a trajectory prediction dataset designed to capture realistic pedestrian movement patterns in confined indoor spaces.  This dataset is more challenging due to the increased complexity of indoor scenes.
\item We provide a robot navigation emulator which is developed to mimic the real-world environments, allowing researches to visually create custom scenes, generate data, and conduct testing.
\item Our dataset supports training and testing for various tasks, including object detection, semantic segmentation, robot relocalization, trajectory prediction, and navigation. We provide benchmarks and analyses of several state-of-the-art methods for each task.
\end{itemize}

\textbf{Novelty with respect to our previous work \cite{tang2024mobile}.} An earlier conference version of THUD++ was presented in \cite{tang2024mobile}, this paper builds upon that work with three significant contributions:
\begin{itemize}
    \item We have integrated the Unity3D-based simulation platform and provided a robot navigation emulator with a user-friendly interface. An efficient evaluation pipeline has also been integrated for both static and dynamic mobile robot indoor tasks, including 3D object detection, semantic segmentation, robot relocalization, trajectory prediction, and navigation. This allows researchers to visually create custom scenes, generate personal data, and perform testing non-invasively in future studies.
    \item We have enriched the dataset by generating a high-quality extension tailored for dynamic pedestrian trajectory prediction in indoor scenes. To the best of our knowledge, this represents one of the largest datasets specifically designed for dynamic pedestrian trajectory prediction in indoor environments.
    \item We conducted additional experiments on dynamic scene tasks using the THUD++ dataset, such as pedestrian trajectory prediction (section \ref{Ped Traj Pred}) and robot navigation (section \ref{Path Planning}), verifying that THUD++ satisfies the training and testing requirements for dynamic scene understanding tasks. Compared to other mobile robot datasets, THUD++ is uniquely well-suited for tasks involving complex dynamic scenes.
\end{itemize}

\section{Related Work}
\subsection{RGB-D Datasets}

\begin{table*}[t]
    \centering
    \caption{RGB-D Datasets Comparison}
    \label{RGB-D Datasets}
    \begin{tabular}{clcclccccccc}
    \toprule
    \multirow{2}*[-0.75ex]{\makecell{\textbf{Type}}} & \multirow{2}*[-0.75ex]{\textbf{Dataset}} & \multirow{2}*[-0.75ex]{\makecell{\textbf{Data}\\\textbf{type}}} & \multirow{2}*[-0.75ex]{\textbf{Year}} & \multirow{2}*[-0.75ex]{\textbf{\# Labels}} & \multirow{2}*[-0.75ex]{\makecell{\textbf{\# Annotations}\\\textbf{per frame}}} & \multirow{2}*[-0.75ex]{\makecell{\textbf{\# Object}\\\textbf{classes}}} & \multirow{2}*[-0.75ex]{\makecell{\textbf{Dynamic}\\\textbf{objects}$^a$}} & \multicolumn{4}{c}{\textbf{Tasks}$^b$} \\
    \cmidrule{9-12} 
    & & & & & & & & \textbf{\textit{2D}} & \textbf{\textit{3D}}& \textbf{\textit{SS}}& \textbf{\textit{RC}} \\
    \midrule
    \multirow{5}{*}{2D} & B3DO \cite{B3DO} & Real & 2011 & 849 frames & 2$\sim$5 & 50+ & × & \checkmark & × & × & × \\
    & NYU-Depth v2 \cite{NYU} & Real & 2012 & 1,449 frames & 30$\sim$40 & 894 & × & × & × & \checkmark & × \\
    & SUN3D \cite{SUN3D} & Real & 2013 & 8 scans & 10$\sim$15 & - & × & × & × & \checkmark & \checkmark \\
    & Stanford 2D-3D-S \cite{Stanford} & Real & 2017 & 70,496 frames & 10$\sim$15 & 13 & × & × & × & \checkmark & × \\
    & SceneNet RGB-D \cite{SceneNet} & Synthetic & 2017 & 5M frames & 20$\sim$30 & 255 & × & × & × & \checkmark & \checkmark \\
    \cmidrule{1-12}
    \multirow{8}{*}{3D} & SUN RGB-D \cite{SUNRGBD} & Real & 2015 & 10k frames & 20$\sim$30 & 800 & × & × & \checkmark & \checkmark & \checkmark \\
    & ScanNet \cite{ScanNet} & Real & 2017 & 2.5M frames & 10$\sim$15 & 21 & × & × & \checkmark & \checkmark & \checkmark \\
    & SUN-CG \cite{SUNCG} & Synthetic & 2017 & 500k frames & 5$\sim$15 & 84 & × & × & × & \checkmark & \checkmark \\
    & Matterport 3D \cite{Matterport3D} & Real & 2017 & 194,400 frames & 5$\sim$15 & 40 & × & × & × & \checkmark & \checkmark \\
    & InteriorNet \cite{InteriorNet} & Synthetic & 2018 & 20M frames & 20$\sim$30 & 158 & × & × & × & \checkmark & \checkmark \\
    & ARKitScenes \cite{ARKitScenes} & Real & 2022 & 5,047 scans & 5$\sim$10 & 17 & × & × & \checkmark & × & × \\
    & ScanNet++ \cite{ScanNet++} & Real & 2023 & 1,858 scans & 20$\sim$30 & 1,000+ & × & × & × & \checkmark & \checkmark \\
    \cmidrule{2-12}
    & \textbf{THUD++(Ours)} & \textbf{Synthetic\&Real} & \textbf{2023} & \textbf{90k frames} & \textbf{150$\sim$200} & \textbf{91} & \checkmark & \checkmark & \checkmark & \checkmark & \checkmark \\
    \bottomrule
    \multicolumn{11}{l}{$^a$Dynamic objects: Includes moving robots, walking people, and people moving with shopping carts.} \\
    \multicolumn{11}{l}{$^b$2D: 2D Object Detection; 3D: 3D Object Detection; SS: Semantic Segmentation; RC: 3D Reconstruction.}
    \vspace{-10pt}
    \end{tabular}
\end{table*}

Scene understanding is a fundamental topic in robotic environment perception, encompassing various common computer vision tasks such as 3D object detection, object classification, semantic segmentation, pose estimation, spatial layout estimation, and CAD model retrieval and alignment, etc. Consequently, numerous RGB-D datasets have been developed to meet the evolving demands of robotic perception algorithms.

Table \ref{RGB-D Datasets} provides a summary of several popular datasets. Based on their annotation types, these RGB-D datasets are broadly classified into two categories: annotations in the 2D domain and annotations in the 3D domain.

Since obtaining accurate and dense 3D annotations is challenging, some works label RGB-D images with 2D annotations, which can indirectly serve as ground truth for 3D tasks. Berkeley 3D Object Dataset \cite{B3DO} has 2D bounding box annotations on RGB-D images, NYU Depth v2 \cite{NYU} includes 2D semantic segmentation from short RGB-D videos with 1449 selected frames tagged, SUN3D \cite{SUN3D} dataset is composed of 415 RGB-D video sequences in 254 scenes. Stanford 2D-3D-Semantics dataset \cite{Stanford} utilizes the iGibson simulation environment to provide large scale virtual scenes containing 2D texture, geometric, and semantic information. SceneNet RGB-D dataset \cite{SceneNet} comprises 5 million images with diverse types of 2D annotations.

3D annotation is challenging, yet some works have still made significant contributions. SUN RGB-D \cite{SUNRGBD} contains 10,335 RGB-D images with dense 2D/3D annotations, including 2D polygons, 3D bounding boxes with accurate object orientation and 3D room layout. ScanNet \cite{ScanNet} includes 1,513 video sequences annotated with 3D camera poses, surface reconstruction, semantic segmentation, and partially aligned CAD models. SUN-CG \cite{SUNCG} provides 45,000 virtual scene layouts and 500,000 rendered images, including single-view RGB, depth maps, and semantic segmentation maps. Matterport 3D \cite{Matterport3D} contains 194,400 RGB-D images for generating panoramas with surface reconstruction, camera positions, and 2D/3D semantic segmentation annotations. InteriorNet \cite{InteriorNet} is entirely rendered in virtual home scenes and comprises 15,000 sequences. ARKitScenes \cite{ARKitScenes} enhances the resolution of ground truth geometry derived from laser scans. ScanNet++ \cite{ScanNet++} is a new dataset that contains 460 high-resolution 3D reconstructions of indoor scenes with dense semantic and instance annotations.

Drawing references from both 2D and 3D annotation works, current datasets lack dynamic objects and pedestrians, which are common in real-world scenarios and present significant challenges for robot-related research. These datasets often suffer from issues such as poor data quality, small annotation volumes, limited label types, noisy annotations, and unrealistic virtual scene layouts. Current research frequently requires testing, training, and deployment on real hardware within virtual scenarios, a need that existing datasets cannot fully address.

\subsection{Pedestrian Trajectory Datasets}

Pedestrian trajectory prediction is essential in mobile robotics and autonomous driving. It involves predicting future pedestrian positions based on their current and past movements, which is vital for navigation, collision avoidance, and interaction planning in dynamic environments.

Table \ref{Trajectory Datasets} summarizes several widely used pedestrian trajectory prediction datasets. Based on their sampling environments, these datasets can be broadly categorized into two types: indoor and outdoor datasets.

Early pedestrian trajectory datasets focused predominantly on outdoor scenes. The UCY \cite{ucy2007} and ETH \cite{eth2009} datasets are two of the most commonly used in this field, encompassing more than 1,500 pedestrian trajectories. These datasets consist of real pedestrian trajectories captured at a frequency of 2.5Hz in public spaces from an overhead perspective. Each scene includes one or more video sequences, covering different periods and pedestrian flows, and features rich multi-person interaction scenarios. The Stanford Drone Dataset (SDD) \cite{sdd2016} provides approximately 19,000 agent trajectories within the Stanford University campus, including 11,200 pedestrian trajectories.

In recent years, several indoor trajectory datasets have been introduced. The ATC Dataset \cite{brvsvcic2013person} captures pedestrian movements in a shopping center, with automated annotations for tracking pedestrian behavior in large public spaces. The L-CAS Dataset \cite{L-CAS2017} records pedestrian movements in a university building from a moving or stationary robot, featuring 935 automatically annotated pedestrian tracks. The THÖR Dataset \cite{THOR2019} was collected in a laboratory environment, comprising over 600 automatically annotated pedestrian tracks.

The movement patterns of pedestrians in indoor scenes are more complex and are typically influenced by spatial constraints and obstacles. However, existing indoor datasets dedicate less attention to the effects of indoor obstacles. While some datasets model static obstacles as intelligent entities with zero velocity, this approach is insufficient to capture the full complexity of indoor scenes. In fact, models trained on these datasets cannot be directly applied to pedestrian prediction tasks in real-world indoor environments.

\begin{table*}[t]
    \centering
    \caption{Pedestrian Trajectory Datasets Comparison}
    \label{Trajectory Datasets}
    \begin{tabular}{clcllclc}
    \toprule
    \multicolumn{1}{l}{\textbf{Environment}} & \textbf{Dataset} & \textbf{Year} & \textbf{Scenes} & \textbf{Duration} & \textbf{\# Scenes} & \textbf{\# Pedestrians} & \textbf{Annotation} \\
    \midrule
    \multirow{3}{*}{Outdoor}  & UCY\cite{ucy2007}     & 2007 & University, Street  & 16.5 min & 2  & 786   & Manual \\
      & ETH\cite{eth2009}  & 2009 & University, Hotel  & 25 min   & 2 & 750 & Manual\\
      & SDD\cite{sdd2016}     & 2016 & University  & 5 hours  & 6 & 11,200     & -  \\
    \cmidrule{1-8}
    \multirow{4}{*}{Indoor}   & ATC\cite{brvsvcic2013person}     & 2013 & Shopping Centre & 92 days  & 1  & -   & Auto\\
      & L-CAS\cite{L-CAS2017}   & 2017 & Minerva Building  & 49 min   & 1   & 935  & Auto\\
      & THÖR\cite{THOR2019}    & 2019 & University Lab   & 60 min   & 1  & over 600  & Auto \\
      & \textbf{THUD++(Ours)}    & \textbf{2024} & \textbf{Supermarket, Office, Gym} & \textbf{60 min}   & \textbf{3}  & \textbf{1257}   & \textbf{Auto} \\
    \bottomrule
    \end{tabular}
\end{table*}

\section{Data Acquisition Platform}

The data acquisition platform consists of two parts: real-world platform and simulation platform. The real-world platform collects data from mobile robots in real environments, while the Unity3D-based simulation platform collects data from virtual sensors in synthetic environments.

\subsection{Real-World Robot Platform}
\begin{figure}[h]
    \centering
    \includegraphics[width=8.5cm]{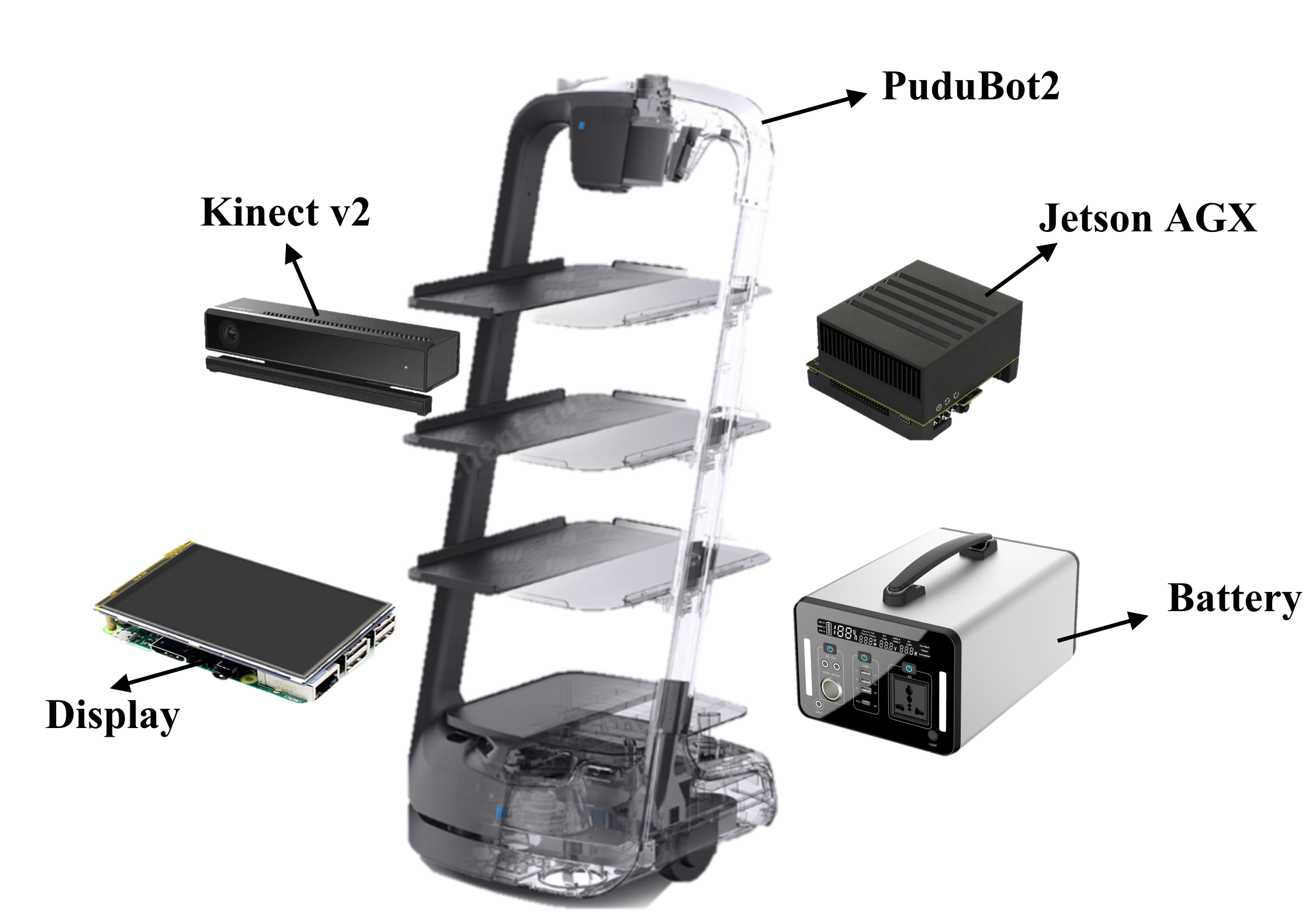}
    \caption{PUDUbot2\&Kinect V2 integrated collection platform}
    \vspace{-5pt}
    \label{fig2}
\end{figure}

Real-world data collection utilizes the PUDUbot2\&Kinect V2 integrated collection platform, as shown in Fig.~\ref{fig2}. PUDUbot2, an advanced mobile service robot, records real-time robot poses (translation distance and Euler angles) while operating in developer mode. Depth and RGB images of real data are collected through Kinect V2.

The real-world data collection system comprises two independent subsystems: pose acquisition and image acquisition. The acquisition of robot pose data primarily relies on the V-SLAM localization at the top of PUDUbot2, supplemented by the odometry of the steering wheel. The data is computed at a frame rate of 40 fps. 

The image data comprises both depth and RGB images. We developed ROS scripts for synchronous acquisition of RGB and depth images. The depth images in the released version have been denoised using the Self-Supervised Deep Depth Denoising method \cite{Sterzentsenko_2019_ICCV}. Due to the inherent instability of the Kinect V2 sensor, the actual frame rate for capturing RGB and depth images fluctuates between 15 and 30 fps.

Additionally, timestamp alignment is necessary. To align the timestamps, we consider the overlapping images of the signal board captured by the PUDUbot2 depth camera and the Kinect V2 as the same moment. However, during actual testing, we observed an alignment error of 0-3 frames using this method. To better match the pose data with the fluctuating image frame rate, we employed cubic spline interpolation. The pose data, initially recorded at 40 fps, was upsampled to 2000 fps using cubic spline interpolation. The interpolated pose timestamps were then matched with the closest pose timestamp to the corresponding image timestamp.

\subsection{Unity3D-based Simulation Platform}
\begin{figure*}[ht]
    \centering
    \includegraphics[width=\linewidth]{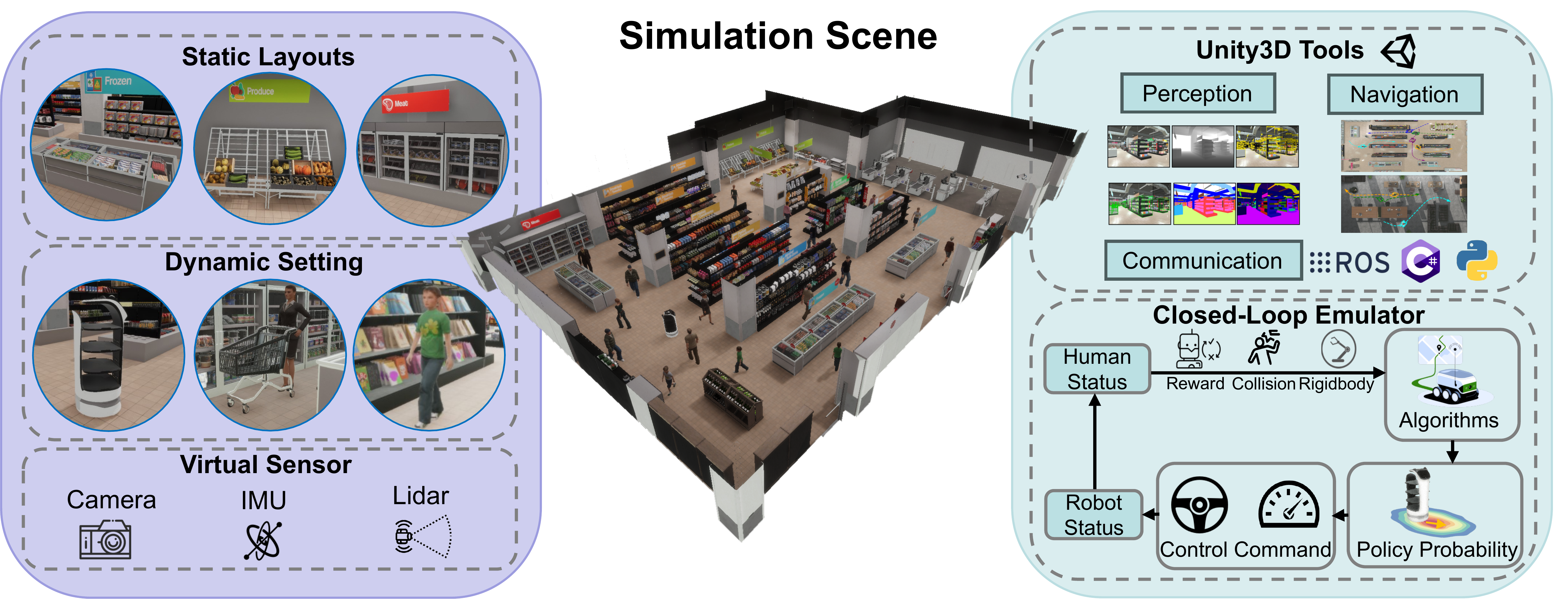}
    \caption{Unity3D-based simulation platform}
    \vspace{-10pt}
    \label{fig1}
\end{figure*}

The Unity3D-based platform is designed to simulate realistic working environments for mobile robots, enabling the collection of virtual sensor data, including RGB images, depth maps, robot pose, and IMU data. The platform has been developed using open-source tools from Unity3D and has undergone secondary development to meet the specific requirements of the synthetic data acquisition for mobile robots. The Unity3D physics engine is utilized to simulate the physical movements of the mobile robot while simultaneously publishing the collected synthetic data and labels via ROS2. The platform architecture is shown in Fig.~\ref{fig1}. The platform ultimately aims to provide a comprehensive solution for synthetic data acquisition, enabling the training and testing of algorithms for mobile robot indoor scene understanding.

The synthetic data is generated by creating virtual sensors in Unity3D scenes. Based on the type and position of sensors on the real mobile robot, the camera for the physical simulation robot is configured accordingly, with a height of 1.2m and a pitch angle of 0 degrees. As the robot navigates through the scenes, \textit{RGB images, depth maps, robot poses}, and \textit{IMU} data are collected. RGB images and depth maps are generated at a resolution of \textit{730×530}, while the robot's pose is described using its \textit{xyz} coordinates and its rotation angle around the \textit{z-axis}. Besides, Unity3D’s built-in navigation tools enable the creation of dynamic environments and the simulation of realistic pedestrian behavior, generating high-fidelity trajectory data for indoor scenes.

Additionally, the robot navigation emulator is further developed based on this Unity3D simulation platform. Compared to the off-line path planning datasets, the on-line simulation platform is more suitable for closed-loop training and testing. By utilizing the A-star algorithm \cite{hart1968formal} and personalized planning algorithm as global and local planning strategies respectively, the ego-robot could navigate purposefully. We provide the personalized algorithms interface for training and testing.

\section{RGB-D Dataset Construction}
\subsection{Data Acquisition}

\subsubsection{Real Data Acquisition}
\textbf{}

Data collection for this dataset is specifically tailored to the field of mobile robotics. To better serve this purpose, real-world data was collected from typical service robot environments on the Tsinghua University campus. As shown in the second column of Fig.\ref{fig0}. These include eight scenes: a corridor of laboratory, the lobby of teaching building, department meeting rooms, departmental visiting laboratories, a cafeteria, a campus dining hall, elevator area, and shops.

\begin{figure}[t]
    \centering
    \includegraphics[width=8.5cm]{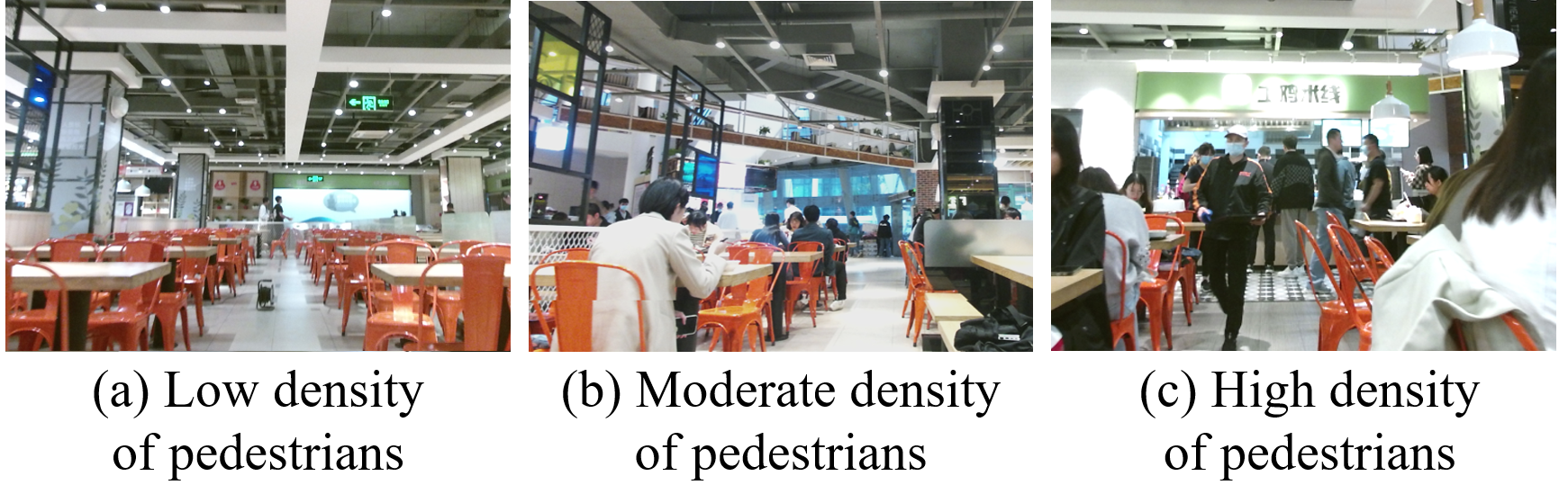}
    \caption{Scenes with varying levels of dynamic complexity}
    \vspace{-8pt}
    \label{fig3}
\end{figure}

One of the distinguishing features of THUD++ is its dynamic nature. To showcase varying levels of dynamic complexity in real-world scenarios, data was collected at different time intervals. For example, Fig.~\ref{fig3} illustrates the data collection process in the canteen scene at different time periods. Within different data sequences in the same scene, there are sequences with low, moderate and high density of pedestrians. These sequences with varying levels of dynamic complexity in the same environment will help users evaluate the algorithm's capabilities and better align with the real-world situations that service robots encounter at different times. Another notable feature of this dataset is the inclusion of various challenging labels, such as elevators, glass objects, and other item categories. These items significantly influence robot perception during real-world testing.

To annotate the real-world data with 2D and 3D object information, we first preprocessed RGB images and depth maps by applying automated algorithms to predict the 2D\&3D positions and bounding boxes of objects, such as Faster-RCNN \cite{FasterRCNN} and Votenet \cite{ImVoteNet}. However, the algorithms for 3D objects detection are not very effective. Subsequently, we performed manual review and verification to correct and refine the annotated bounding boxes, ensuring the accuracy and completeness of the detected objects. For the 2D semantic and instance segmentation task, we similarly employed a semi-automated annotation method, manual review and correction were also conducted by annotators to ensure accurate pixel-wise classification and semantic labeling.

\subsubsection{Synthetic Data Acquisition}
\textbf{}

\begin{figure}[t]
    \centering
    \includegraphics[width=\linewidth]{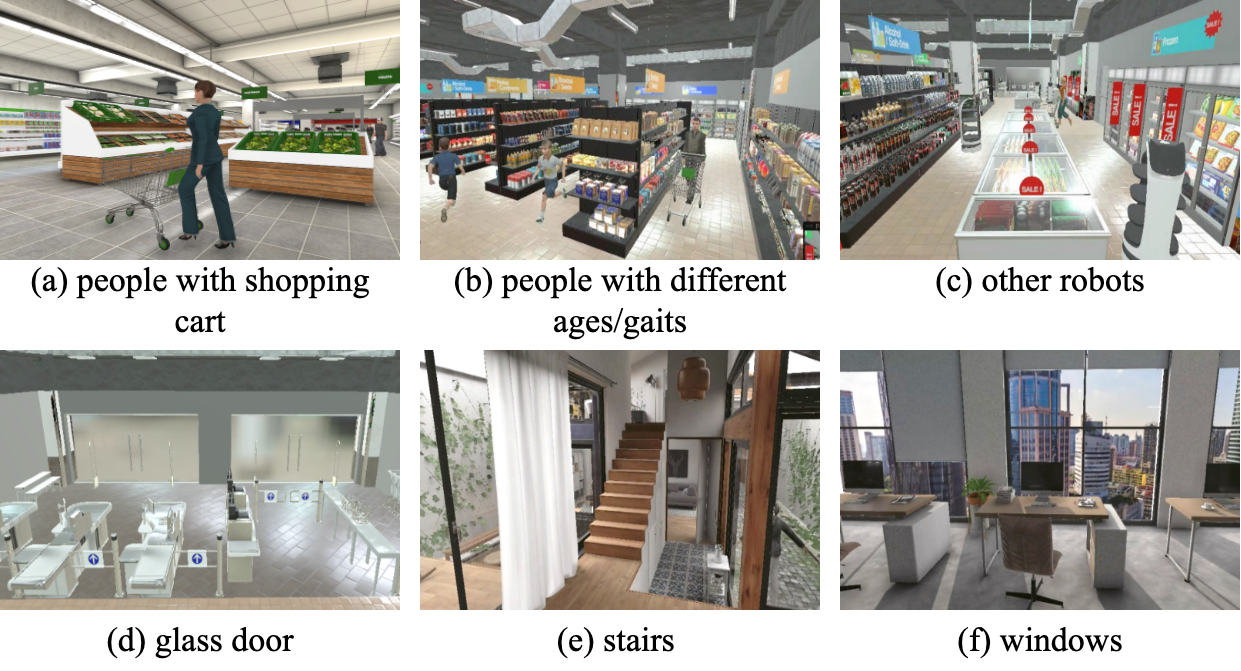}
    \caption{Dynamic and special simulation scenarios}
    \vspace{-8pt}
    \label{fig:scenes}
\end{figure}

To acquire more realistic and effective data, the scenes were designed based on the actual working environments of mobile robots, taking into account two main aspects: dynamic scenarios with moving obstacles and special scenarios that pose potential hazards for robots in the reality. As illustrated in Fig.~\ref{fig:scenes}. Dynamic scenarios were created by setting up pedestrians in the scenes, some special cases were also considered, such as running children, people pushing shopping carts, and other moving robots in the scene. The scenes also take into account pedestrians of varying ages and different clothing styles, including those of men, women, and children. For special scenarios, the scene settings focus on realistic objects that tend to pose difficulties for robotic tasks, such as stairs, windows, glass doors, etc.

The annotations for the simulated data were automatically generated using our custom-built data acquisition platform. We integrated the robot model of \textit{Bella} into the virtual scenes for data collection, allowing the robot to navigate and capture data from its own perspective. During the data collection process, we simultaneously extracted vital information, including 2D and 3D bounding boxes, semantic and instance segmentation images, robot's pose, and IMU information. These annotations were then synchronized with the captured RGB images and depth maps, ensuring a one-to-one correspondence. Sequentially, the annotation information extracted from our data acquisition platform serves as ground truth within the virtual scenes.

\subsection{Data Statistics}
Our dataset comprises a total of 90,175 annotated frames, consisting of 84,984 frames from synthetic data collection and 5,191 frames from real-world data collection. Each frame has undergone intensive annotation, resulting in a total of over 20M labels, with over 1.2M labels for dynamic objects such as pedestrians, robots, and shopping carts. On average, each frame contains 176 data labels.

\begin{figure*}[t]
    \centering
    \includegraphics[width=\linewidth]{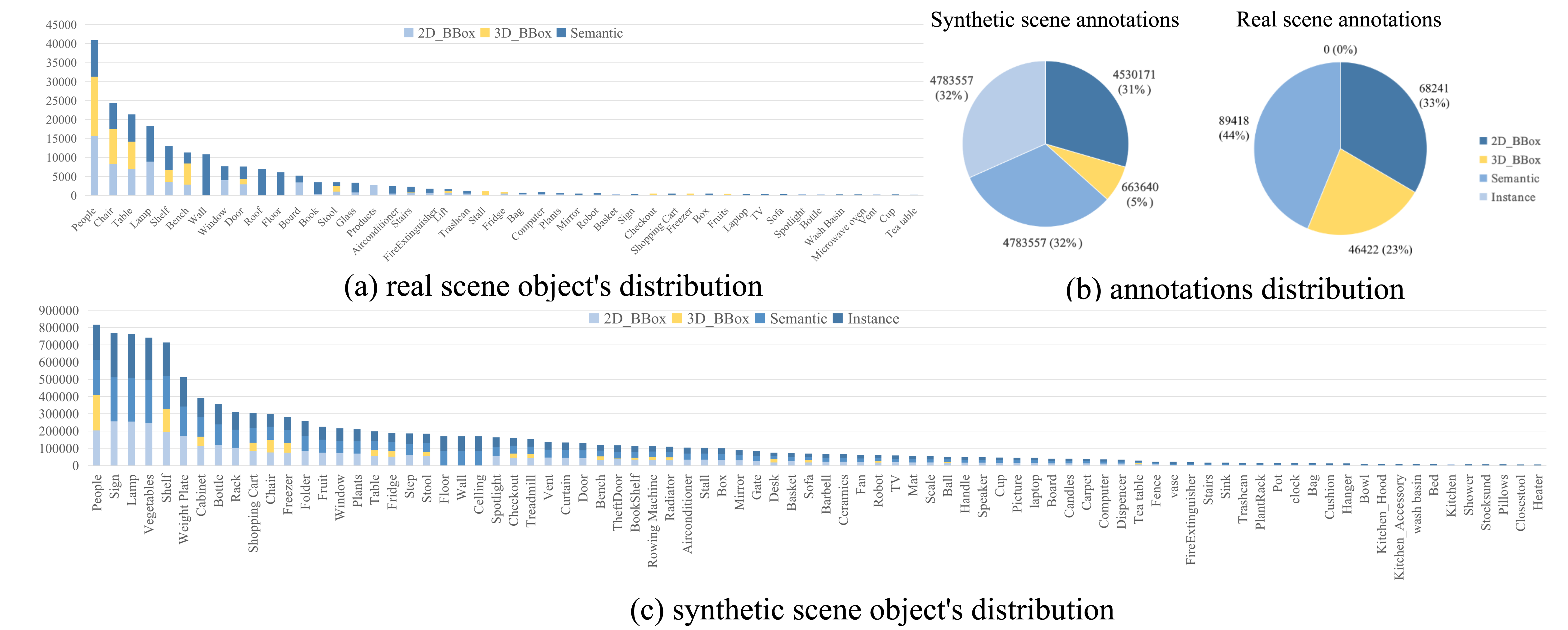}
    \caption{Statistics of annotations in our dataset}
    \vspace{-8pt}
    \label{annotaion1}
\end{figure*}

These labels encompass four annotation types: 2D bounding boxes, 3D bounding boxes, semantic segmentation, and instance segmentation, with their respective proportions as shown in Fig.~\ref{annotaion1}(b). The dataset encompasses 8 real and 5 synthetic large-scale indoor scenes, with average area exceeding $300m^2$. There are 91 different object categories within these scenes, and their distribution, along with the counts of each annotation type, is illustrated in Fig.~\ref{annotaion1}(a)(c).

\section{Trajectory Dataset Construction}

\subsection{Data Acquisition}

\textbf{\textit{Acquisition Scenarios:}} 
Similar to the acquisition of RGB-D data, the scenes captured cover a wide range of indoor environments designed to reflect real-world environments, including furniture, walls and other obstacles. These scenes are designed to capture realistic pedestrian movement patterns in confined spaces. The dimensions are comparable to real-world environments, such as 13m × 28m for the Gym, 7.8m × 16.2m for the Office, and 19m × 32m for the Supermarket. Dynamic elements, such as moving pedestrians, have been incorporated to simulate typical indoor pedestrian traffic. The scenes were carefully designed to include a range of layouts, ensuring the completeness of the dataset. Furthermore, in order to ascertain the variability of the synthetic data in comparison to the real data, an open scene without obstacles was set up in order to simulate an outdoor environment. This configuration allows for a comparison of experimental results with those from publicly available datasets.

\textbf{\textit{Acquisition Method:}} 
The synthetic data was generated by exporting the coordinate data of pedestrians in Unity3D. For each scene, a parameter randomisation strategy was employed to generate different sequences of data. Parameters such as the number of pedestrians, their initial positions, and the time intervals between their appearances can be configured. Similarly, the trajectory of each pedestrian can be defined by their starting and ending points, direction, speed, angular velocity, and acceleration. Furthermore, each pedestrian will cease to exist immediately upon reaching the target point, thus eliminating the possibility of a pedestrian remaining in place, which would otherwise result in the generation of invalid data. By adjusting these parameters and recording at 0.4 second intervals, a variety of trajectory data sequences were generated in different scenarios.

\subsection{Data Statistics}
The data is exported in a structured format to facilitate subsequent analysis and model training. The frame IDs in the data commence at 0 and are enumerated in increments of 10. The identification numbers for pedestrians commence at 1 and are incremented by 10. Each pedestrian in each sequence is assigned a unique identification number. The \textit{x} and \textit{y} coordinates are recorded in the world coordinate system, and the origin can be defined as required. Each trajectory sequence is saved as a text file with a format comprising a frame ID, a pedestrian ID, and two coordinates. 

The dataset includes three indoor scenes: Gym, Office, and Supermarket, each with a different set of pedestrian trajectories. The dataset contains 6363 frames, with 1257 automatically annotated pedestrian tracks and an average of 5.9 pedestrians per frame.

\section{Evaluation on THUD++ Dataset}

To verify whether our dynamic RGB-D dataset is necessary and useful for mobile robots, we evaluate three tasks: 3D object detection, semantic segmentation, and robot relocalization. Furthermore, we applied the trajectory dataset to pedestrian trajectory prediction and evaluated the robot navigation algorithm using the robot navigation emulator. Important results and conclusion are presented here, with a detailed analysis available on the \textit{project website}.

\begin{figure*}[t]
    \centering
    \includegraphics[width=16cm, height=5.5cm]{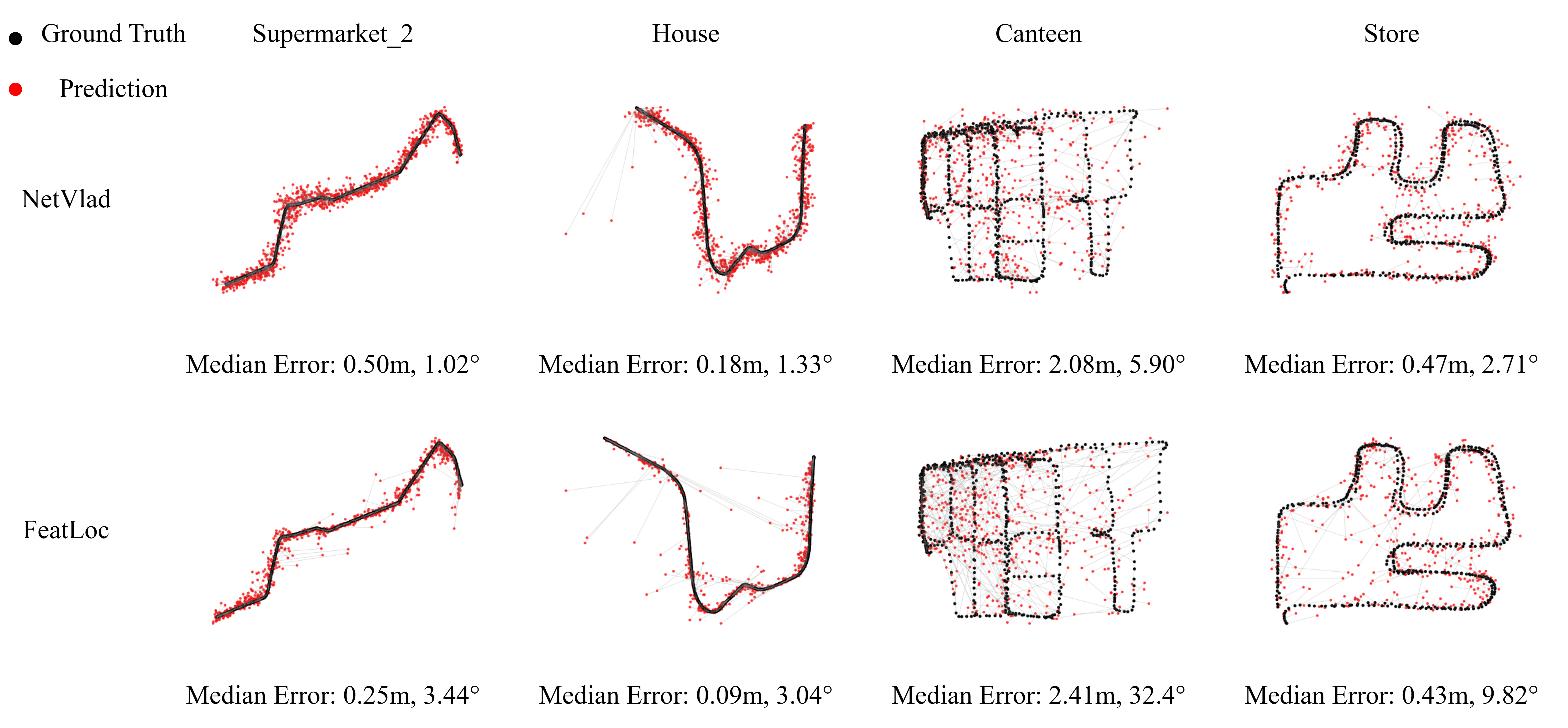}
    \caption{Test result visualization for robot relocalization}
    \label{fig_visualization}
    \vspace{-10pt}
\end{figure*}

\begin{table}[t]
\centering
\caption{Experiments on 3D Object Detection}
\label{tab:3D object detection results}
\begin{tabular}{cccc}
    \toprule
    \multirow{2}{*}{\textbf{Scene}}  & \multirow{2}{*}{\textbf{Method}} & \textbf{Dynamic Objs} & \textbf{Static Objs}   \\
     & & \textbf{(mAP)} & \textbf{(mAP)} \\
    \midrule
    \multirow{3}{*}{Supermarket} & F-PointNet & 7.89 & 8.92\\& ImVoteNet  & 17.49 & 17.29 \\& DeMF  & 34.51 & 38.24\\
    \midrule
    \multirow{3}{*}{Canteen} & F-PointNet & 18.16 & 36.67\\& ImVoteNet  & 26.72 & 43.37 \\& DeMF  & 28.56 & 45.43\\
    \bottomrule
    \vspace{-10pt}
\end{tabular}
\end{table}

\subsection{3D Object Detection}
3D object detection methods estimate a 3D bounding box and a pose of each object presented in a scene \cite{DeformableDETR, wang_tokenfusionmultimodal_2022, wang_frustum_2019}. To compare the performance difference, three representative indoor 3D object detection algorithms which have good performance on both dynamic and static objects, i.e. F-PointNet \cite{F-PointNet}, ImVoteNet \cite{ImVoteNet} and DeMF \cite{DeMF}, are selected. These algorithms were evaluated on the real-world canteen scene and the synthetic supermarket scene to assess their mean Average Precision (mAP) for static and dynamic objects. The experimental results are depicted in Table \ref{tab:3D object detection results}. In the supermarket scene, the selected 5 static objects include \textit{Chair, Table, Shelf, Cabinet, Fridge}. While the 3 dynamic objects encompass \textit{Shopping cart, Robot, People}. In the canteen scene, the 7 static objects are \textit{Chair, Table, Shelf, Bench, Door, Stairs, Stool}, and the dynamic object is only \textit{People}. Moreover, we used the average number of dynamic objects per frame as a metric to assess the level of scene dynamic complexity, the supermarket and canteen scene dynamic complexity is 0.94 and 3.34, respectively.

We can see two important conclusion from the test results in Table \ref{tab:3D object detection results}. First, the results indicate that different algorithms have varying degrees of decrease in accuracy when detecting dynamic objects as compared to static objects. Second, 
it can be observed that having a larger density of dynamic objects lead to more significant accuracy difference between static and dynamic objects. But this could also be influenced by the object categories.


\subsection{Semantic Segmentation}
Semantic segmentation methods aim to assign each pixel in an image to its corresponding semantic category, focusing on the geometric structure and differentiation of various objects in the scene \cite{SS_survey, SS_survey_2, SS_survey_3}. However, this task is particularly challenging in the context of robotic applications, where indoor environments introduce several complexities, such as diverse object shapes and sizes, occlusions, lighting variations, and the inherent clutter in real-world settings. Four RGB-D semantic segmentation algorithms, i.e. ACNet \cite{ACNet}, RedNet \cite{RedNet}, ESANet \cite{ESANet}, and SA-Gate \cite{SA-Gate}, are selected and tested on the aforementioned synthetic supermarket scene and real-world canteen scene. 31 and 19 static and dynamic objects with semantic labels are added to the supermarket and canteen scene, respectively. Objects without labels are set to the void value with corresponding pixels (0, 0, 0).


\begin{table}[t]
\caption{Experiments on Semantic Segmentation}
\vspace{-10pt}
\begin{center}
\label{tab:SS_results}
\begin{tabular}{cccc}
    \toprule
    \textbf{Scene} & \textbf{Method} & \textbf{Backbone} & \textbf{MIoU(\%)}  \\
    \midrule
    \multirow{4}{*}{Supermarket} & ACNet & 3×R50 & 74.83 \\
     & RedNet  & 2×R34 & 76.92 \\
     & ESANet & 2×R34 & 78.42 \\
     & SA-Gate & 2×R101 & 83.19 \\
    \midrule
    \multirow{4}{*}{Canteen} & ACNett & 3×R50 & 51.85 \\
     & RedNet & 2×R34 & 59.83 \\
     & ESANet & 2×R34 & 65.97 \\
     & SA-Gate & 2×R101 & 58.34 \\
    \bottomrule
    \vspace{-15pt}
\end{tabular}
\end{center}
\end{table}

\begin{figure}[b]
    \centering
    \includegraphics[width=\linewidth]{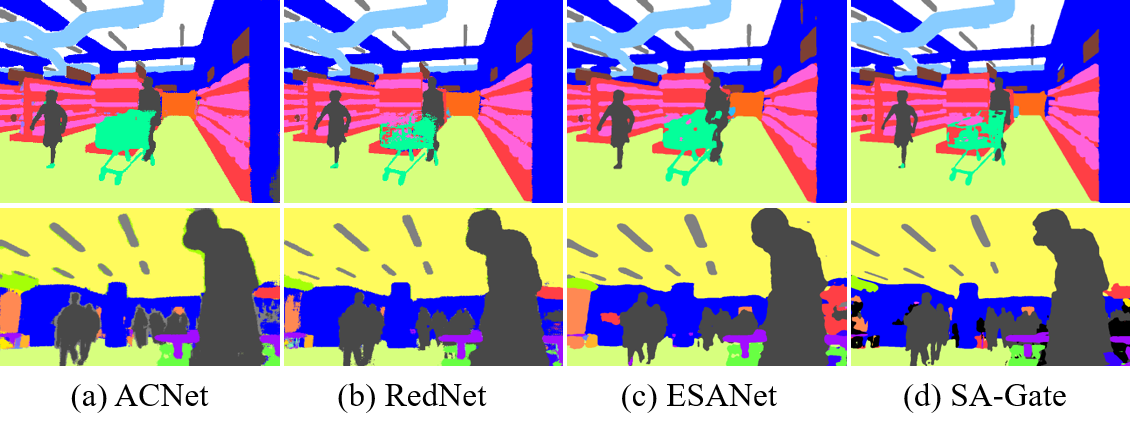}
    \caption{Comparison of semantic segmentation methods}
    \vspace{-10pt}
    \label{fig:SS_results}
\end{figure}

The accuracy of different semantic segmentation methods is measured by calculating the Mean Intersection over Union (MIoU) metric during testing and the results are shown in Table \ref{tab:SS_results}. We can obtain three conclusions. First, the segmentation accuracy on the real-world canteen scene is significantly lower than on the synthetic supermarket scene. Second, the MIoU results for the tested methods are comparable to their original published results, though tested on different datasets. Third, to obtain a quantitative comparison on dynamic objects, we also train and test the ESANet \cite{ESANet} algorithm in the supermarket scene with and without dynamic objects, the MIoU of the test is 78.42\% and 79.63\% respectively. There is no significant difference, which is coincident with our intuition. 
In addition, as shown in Fig.~\ref{fig:SS_results}, different methods still have varying segmentation ability on dynamic object, such as the people and shopping carts.


\subsection{Robot Relocalization}
Robot relocalization refers to the process of accurately determines and adjusts its position based on the environmental perception information and its own localization capabilities, during its movement \cite{posenet, torii2013visual,dfnet}. This process is crucial for enabling robots to navigate and reorient themselves in complex and dynamic environments. Existing methods can be coarsely classified into global- and local-feature based methods. 

We conducted extensive experiments on THUD++ using a representative global- and local-feature based methods, i.e. NetVlad \cite{NetVLAD} and FeatLoc \cite{FeatLoc}, respectively. Two sequences were randomly selected for testing, and the results on both synthetic and real-world scene are visualized in Fig.~\ref{fig_visualization}. 

\begin{figure}[b]
    \centering
    \includegraphics[width=\linewidth]{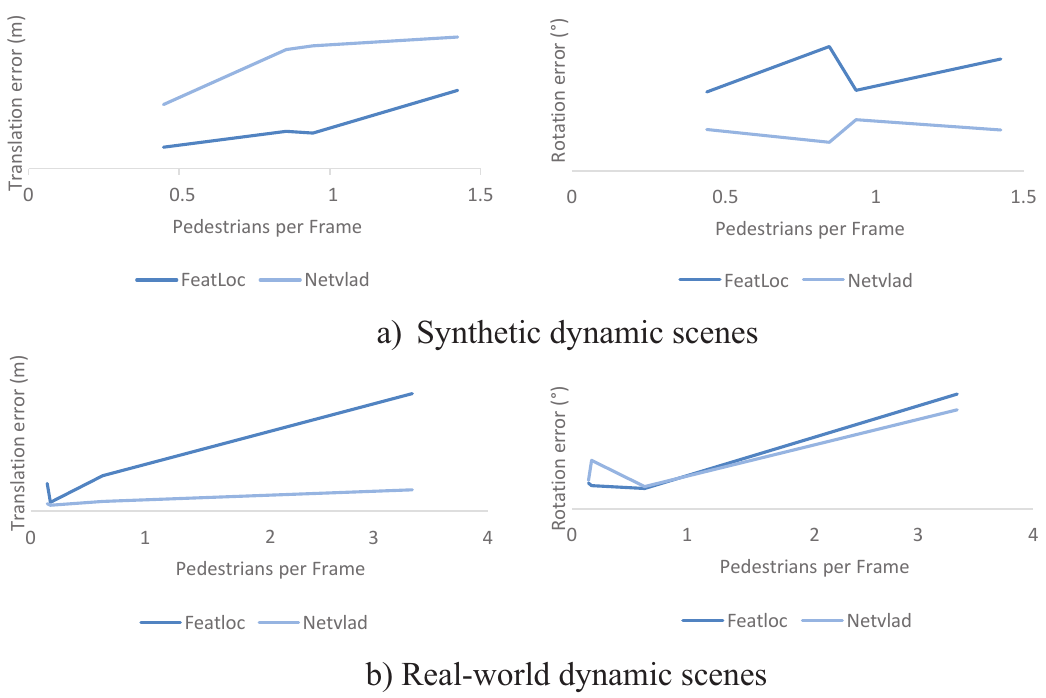}
    \caption{Trans.\& rot. error with different dynamic complexity}
    \label{fig_relocalization_dynamic}
    \vspace{-0.3cm}
\end{figure}

In addition, to analyze the effect of the number of dynamic objects (primarily pedestrians) in various scene images, we define a metric to represent a scene's dynamic complexity, computed as the average number of dynamic pedestrians per frame across the entire dataset. As shown in Fig.~\ref{fig_relocalization_dynamic}, with the scene's dynamic complexity increasing, the accuracy of both methods tends to decrease. Notably, FeatLoc \cite{FeatLoc} exhibits a more significant accuracy drop compared to NetVlad \cite{NetVLAD}, which can be attributed to dynamic objects introducing disturbances and noise during the extraction of scene image features. Moreover, local-feature based methods are more sensitive to dynamic changes in the environment, while global-feature based methods, which rely on broader contextual understanding, tend to be more resilient to such variations.

\subsection{Pedestrian Trajectory Prediction}\label{Ped Traj Pred}

\begin{table}[b]
\centering
\caption{Experiments on Pedestrian Trajectory Prediction. Gym, Office and Supermarket are our synthetic indoor scenes, while ETH \cite{eth2009} is the official public outdoor dataset.}
\label{tab: ped traj predict}
\begin{tabular}{cccc}
    \toprule
    \textbf{Scene} & \textbf{Method} & \textbf{ADE} $\downarrow$ & \textbf{FDE} $\downarrow$ \\ \midrule
    \multirow{3}{*}{Gym} & Social-GAN & 1.47 & 2.98 \\
     & PECNet & 1.69 & 3.06 \\
     & Social-STGCNN & 1.18 & 2.17 \\ \midrule
    \multirow{3}{*}{Office} & Social-GAN & 1.68 & 3.45 \\
     & PECNet & 1.94 & 3.49 \\
     & Social-STGCNN & 1.36 & 2.53 \\ \midrule
    \multirow{3}{*}{Supermarket} & Social-GAN & 1.81 & 3.78 \\
     & PECNet & 1.88 & 3.61 \\
     & Social-STGCNN & 1.55 & 2.91 \\ \midrule
    \multirow{3}{*}{ETH} & Social-GAN & 1.08 & 2.14 \\
     & PECNet & 0.91 & 1.56 \\
     & Social-STGCNN & 0.74 & 1.48 \\ \bottomrule
    \vspace{-10pt}
\end{tabular}
\end{table}

\begin{figure}[h]
    \centering
    \includegraphics[width=\linewidth]{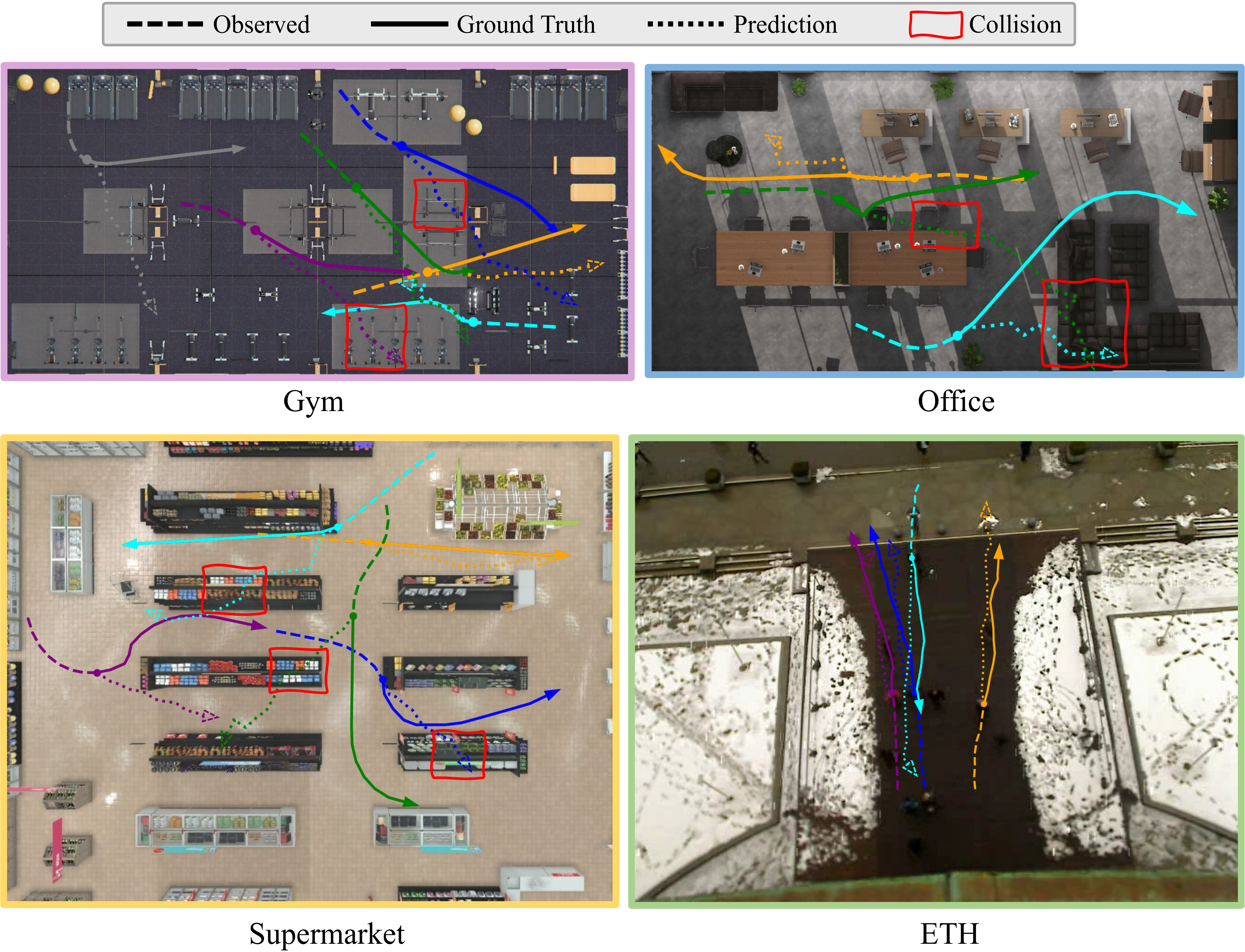}
    \caption{The visualization for pedestrian trajectory prediction. Each color denotes a pedestrian. The closer the predicted trajectory (short dashed line) approximates GT (solid line), the higher the prediction accuracy. Due to the density of static obstacles indoors, the predicted future trajectory may lead to collisions (red rectangular box).}
    \label{fig_trajectory}
    \vspace{-12pt}
\end{figure}

Pedestrian trajectory prediction aims to forecast the future path of a target individual by leveraging their past trajectory and surrounding context. Modeling human movement patterns and human-human interactions is essential to predict socially feasible trajectories.
Three trajectory prediction algorithms, i.e. Social-GAN \cite{socialgan2018}, Social-STGCNN \cite{stgcnn2020} and PECNet \cite{pecnet2020}, are tested on our synthetic indoor scenes (Gym, Office and Supermarket) as well as the official public outdoor dataset (ETH \cite{eth2009}). We choose the ADE (Average Displacement Error) and FDE (Final Displacement Error) as metrics to evaluate the prediction performance, and the lower ADE and FDE indicate the better performance. The experimental results are depicted in Table \ref{tab: ped traj predict}.  In addition, to analyze the moving patterns and collision avoidance strategies of pedestrians, we sample four scenarios and visualize the predicted trajectory in Fig.~\ref{fig_trajectory}. 

In general, there is a large performance decrease for all three methods when it comes to the indoor scenes compared with outdoor ETH scene. For instance, the ADE metric for Social-GAN drops from 1.08 to 1.81 (67.6$\%$) when generalizing from ETH scene to the Supermarket scene. We analyze this performance drop from two aspects. First, in contrast to the outdoor scene, narrow indoor spaces are often filled with a large number of static obstacles, which may interfere with human trajectory decision-making and lead to collisions, as shown in Fig.~\ref{fig_trajectory}. Second, the indoor human interactions occur more frequently because of communication with each other or standing in the way, which makes it more challenging to predict. Overall, we hope that these indoor scenes will facilitate the innovation for pedestrian trajectory prediction.

\begin{table*}[t]
\centering
\caption{Experiments on Robot Navigation. Gym and Supermarket are our synthetic indoor scenes. K represents the number of pedestrians.}
\label{tab: crowdnav different number of people}
\begin{tabular}{cc|ccc|ccc|ccc}
    \toprule
    \textbf{Scene} &  & \textbf{ORCA}  & \textbf{DS-RNN} & \textbf{AttnGraph} & \textbf{ORCA}  & \textbf{DS-RNN} & \textbf{AttnGraph} & \textbf{ORCA}  & \textbf{DS-RNN} & \textbf{AttnGraph}  \\ \midrule
    \multicolumn{2}{c|}{} & \multicolumn{3}{c|}{K=10} & \multicolumn{3}{c|}{K=15} & \multicolumn{3}{c}{K=20}\\ \midrule

    \multirow{4}{*}{Gym} & SR $\uparrow$ & 0.76 & 0.79 & 0.72 & 0.74 & 0.77 & 0.69 & 0.75 & 0.73 & 0.64 \\
     & CR $\downarrow$ & 0.03 & 0.03 & 0.02 & 0.05 & 0.02 & 0.03 & 0.06 & 0.05 & 0.06 \\
     & ANT $\downarrow$ & 22.83 & 13.82 & 14.61 & 27.80 & 16.17 & 15.84 & 30.53 & 14.72 & 18.56 \\
     & SPE & 1.10 & 1.06 & 1.22 & 1.24 & 1.08 & 1.28 & 1.35 & 1.12 & 1.36 \\ \midrule
    
    \multirow{4}{*}{Supermarket} & SR $\uparrow$ & 0.37 & 0.44 & 0.46 & 0.32 & 0.47 & 0.42 & 0.30 & 0.43 & 0.35 \\
     & CR $\downarrow$ & 0.01 & 0.09 & 0.06 & 0.02 & 0.11 & 0.04 & 0.03 & 0.12 & 0.07 \\
     & ANT $\downarrow$ & 37.16 & 25.56 & 30.31 & 39.99 & 27.63 & 32.52 & 46.00 & 27.37 & 36.60 \\
     & SPE & 1.06 & 1.04 & 1.30 & 1.08 & 1.05 & 1.34 & 1.12 & 1.07 & 1.39 \\ \bottomrule
    \vspace{-10pt}
\end{tabular}
\end{table*}

\subsection{Robot Navigation}\label{Path Planning}
Robot navigation involves enabling an autonomous robot to plan and execute a path from its current position to a target location, while avoiding obstacles and satisfying specific constraints. The complexity of robot navigation stems from factors like the dynamic nature of environments.

We conducted extensive experiments in the navigation emulator of THUD++, utilizing three representative local crowd-navigation algorithms: reaction-based method ORCA \cite{van2011reciprocal}, learning-based method DS-RNN \cite{liu2021decentralized}, trajectory prediction-based method AttnGraph \cite{liu2023intention}, each combined with the global A-star algorithm to enable effective long-range navigation.

The simulation environment for training DS-RNN and AttnGraph is designed to replicate real-world dense and dynamic crowd navigation scenarios. It features a 12m×12m 2D plane where the robot interacts with up to 20 humans. The humans, controlled by the ORCA algorithm, have randomized maximum speeds ranging from 0.5 to 1.5m/s and radii between 0.3 and 0.5m. The robot operates with a limited circular sensor range of 5m, observing only humans within this range. It has a continuous action space and a maximum speed of 1m/s. Test environments in THUD++ are configured similarly but incorporate additional static obstacles of varying complexity to simulate more challenging real-world conditions. 

We evaluated all methods on 500 randomly selected unseen cases from THUD++ synthetic indoor scenes. The evaluation metrics are categorized into navigation metrics and social metrics. Navigation metrics assess the Success Rate (SR) and the Average Navigation Time (ANT) in seconds for successful episodes. Social metrics evaluate the robot's social awareness, including the Collision Rate (CR) with other humans and Social Path Efficiency (SPE) which  accounts for the robot's ability to follow efficient paths while being socially aware. We define SPE as the ratio of the robot's average actual path length to the expected path length calculated using A-star algorithm for successful episodes. If the robot runs for more than two minutes without reaching the target, the sample is considered timed out and marked as a failure. 

\begin{figure}[t]
    \centering
    \includegraphics[width=\linewidth]{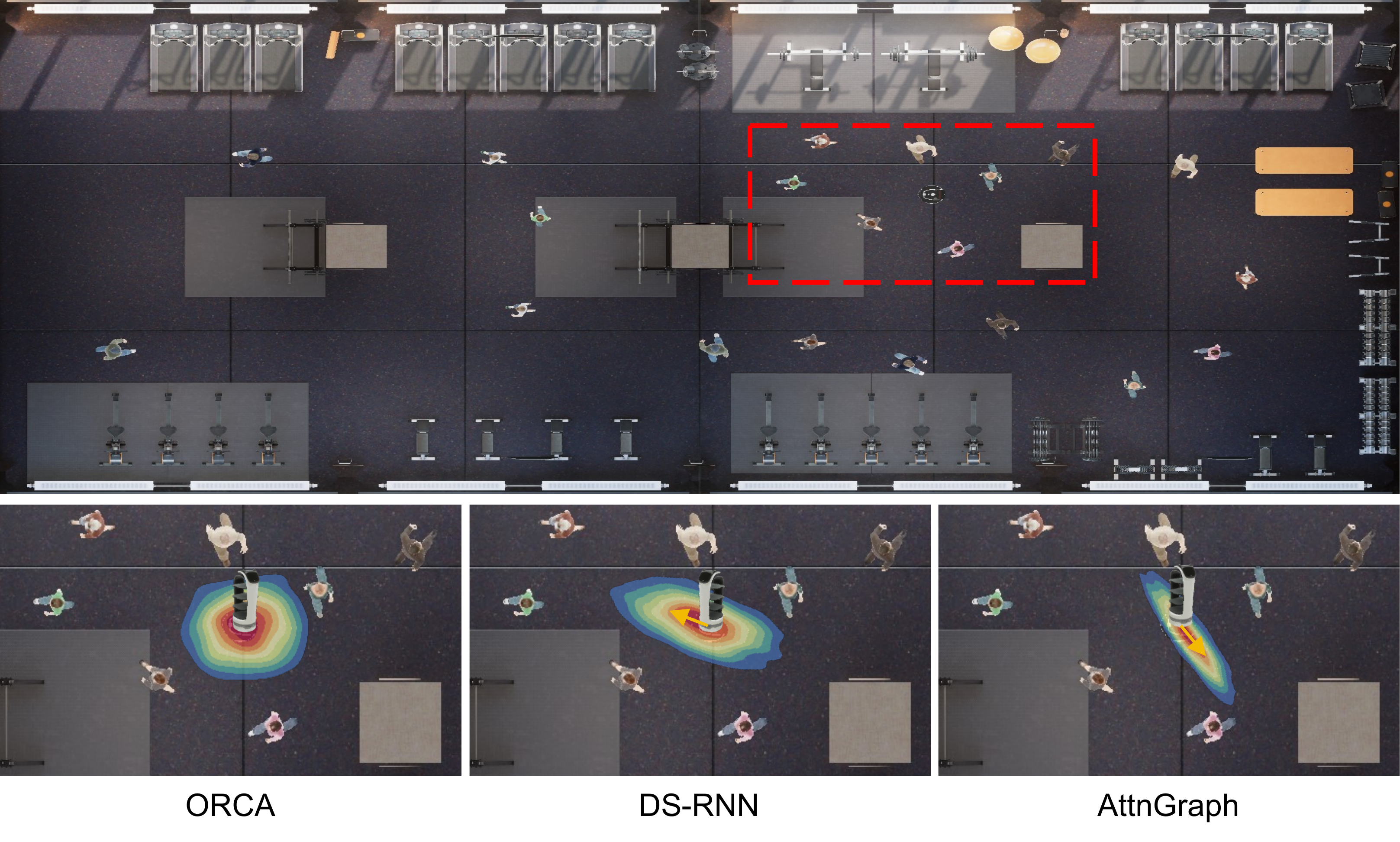}
    \caption{Visualization of freezing robot scenario. The top row displays a typical dynamic indoor scene, while the bottom row shows the prediction probability density for three different algorithms.}
    \label{freezing_robot}
    \vspace{-12pt}
\end{figure}

The experimental results are presented in Table \ref{tab: crowdnav different number of people}. With the number of pedestrians increasing, all methods show performance degradation. SR decreases to varying degrees, while CR increases accordingly. However, ORCA and DS-RNN demonstrate better scalability. AttnGraph’s SR declines significantly as dynamic complexity increases. Specifically, in Gym scene, AttnGraph’s SR decreases from 0.72 to 0.64 (11.1$\%$), while in Supermarket scene, it drops from 0.46 to 0.35 (23.9$\%$). This decline and suboptimal performance may be attributed to the confined nature of spaces like Gym, where increasing dynamic complexity leads trajectory prediction-based methods to render significant portions of the space untraversable. As a result, the robot tends to adopt overly conservative behaviors \cite{driggs2018robust}. Regarding ANT, ORCA exhibits the most significant growth with increasing dynamic complexity. Specifically, in Gym scene, ORCA’s ANT rises from 22.83 to 30.53 (33.7$\%$), while in Supermarket scene, it increases from 37.16 to 46.00 (23.7$\%$). This increase may be attributed to ORCA experiencing \textit{the freezing robot problem} \cite{trautman2010unfreezing} under higher dynamic complexity, as depicted in Fig.~\ref{freezing_robot}. As for SPE, all methods increase with dynamic complexity and show a greater increase in smaller scenes. Specifically, as the number of pedestrians increases, the average increase in SPE for all methods is 12.9$\%$ in Gym scene, whereas in Supermarket scene, it is 4.9$\%$. This indicates that these methods exhibit good scalability in social awareness during navigation.


\section{Conclusions}
In this paper, we introduced the THUD++ (TsingHua University Dynamic) dataset, a mobile robot oriented large-scale indoor dataset for dynamic scene understanding tasks. By incorporating both real-world and synthetic data, THUD++ addresses critical gaps in existing datasets and highlights significant performance challenges in dynamic scenarios.

Extensive evaluations across diverse tasks demonstrate THUD++'s versatility as a benchmarking platform and its suitability for complex dynamic indoor scenes. Furthermore, the dataset’s rich annotations and customizable Unity3D-based simulation platform empower researchers to accelerate the development of innovative robotic algorithms.



\bibliographystyle{IEEEtran}


\vfill

\end{document}